# UTD-Yolov5: A Real-time Underwater Targets Detection Method based on Attention Improved YOLOv5

Jingyao Wang, Naigong Yu

*Abstract*—As the treasure house of nature, the ocean contains abundant resources. But the coral reefs, which are crucial to the sustainable development of marine life, are facing a huge crisis because of the existence of COTS and other organisms. The protection of society through manual labor is limited and inefficient. The unpredictable nature of the marine environment also makes manual operations risky. The use of robots for underwater operations has become a trend. However, the underwater image acquisition has defects such as weak light, low resolution, and many interferences, while the existing target detection algorithms are not effective. Based on this, we propose an underwater target detection algorithm based on Attention Improved YOLOv5, called UTD-Yolov5. It can quickly and efficiently detect COTS, which in turn provides a prerequisite for complex underwater operations. We adjusted the original network architecture of YOLOv5 in multiple stages, including: replacing the original Backbone with a two-stage cascaded CSP (CSP2); introducing the visual channel attention mechanism module SE; designing random anchor box similarity calculation method etc. These operations enable UTD-Yolov5 to detect more flexibly and capture features more accurately. In order to make the network more efficient, we also propose optimization methods such as WBF and iterative refinement mechanism. This paper conducts a lot of experiments based on the CSIRO dataset [1]. The results show that the average accuracy of our UTD-Yolov5 reaches 78.54%, which is a great improvement compared to the baseline.

## I. INTRODUCTION

The ocean is the lifeblood of nature. Among them, coral reefs are critical to marine food webs, habitat supply and nutrient cycling. But now it is facing a huge crisis. Surveys have found that 3/2 of the world's coral reefs are at risk of severe damage or further degradation [2]. In response to this situation, Australian scientists have launched an investigation based on the largest coral reef, the Great Barrier Reef [3]. They found that the main reason for its biological threat is the overpopulation of a particular species of sea star, the coral-feeding crown of thorns starfish (or COTS for short) [4]. Therefore, the society established a large-scale intervention program to control COTS outbreaks and maintain ecological sustainability.

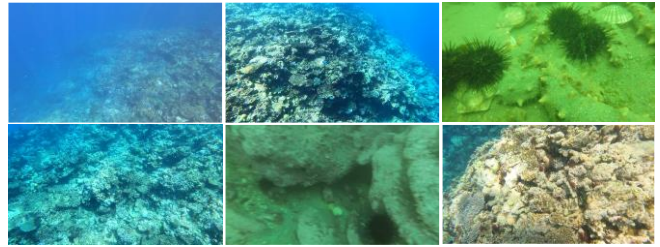

Figure 1. Underwater Images. Compared with normal acquisition, underwater images have problems such as high turbidity, low resolution etc.

The protection of marine biological threats is mostly through artificial fishing to control quantity [5]. However, the number of organisms like CONTS is too large, and it cost a lot, but the efficiency is too low. In addition, due to the risk of underwater operations, there is a greater safety hazard in the artificial fishing of harmful organisms. Therefore, the use of underwater robots has gradually become a trend. The operation of underwater robots can not only reduce the cost of tasks, but also provide data support for scientific research. In order to meet the above expectations, an efficient and real-time underwater target detection algorithm is very important.

With the development of deep learning, the related applications of object detection have gradually matured. This field is mainly divided into two-stage algorithm represented by Faster RCNN [6] and one-stage algorithm represented by YOLOv3 [7]. Among them, SNIPER [8], CornerNet [9], ExtremeNet [10], TridentNet [11], FCOS [12] etc. have achieved the performance of SOTA in many different fields. However, due to the defects of weak light, low resolution, high noise, and overexposure [13] in underwater collected images, the effect of the above algorithm framework is difficult to achieve expectations. The application of Fast RCNN and SGD fusion network [14], cyclic pooling convolution [15], SPP [16], LBP [17] and other algorithms can eliminate the influence of underwater image defects, but only for simple scenes. For coral reefs, the high similarity of CONTS to the surrounding environment makes the detection and segmentation of this species difficult.

We propose an underwater target detection algorithm based on Attention Improved YOLOv5, called UTD-Yolov5. We realize the extraction of high-order features of coral reefs and CONTS by introducing CSP2 and self-attention mechanism through a multi-stage cascade design of the framework. In addition, we design mechanisms such as iterative refinement to adaptively and dynamically find the location of CONTS. In summary, our contributions are mainly as follows:

This work is supported by National Nature Science Foundation under Grant 62076014, project of the Ministry of Industry and Information Technology of China (Z135060009002).

J. Wang and N. Yu are with the Beijing Key Lab of the Computational Intelligence and Intelligent System, Beijing University of Technology, Beijing, China (corresponding author to provide phone: 86-15617877831; e-mail: yunaigong@bjut.edu.cn).

Correspondence: N. Yu, yunaigong@bjut.edu.cn.

- We highly improve YOLOv5 by Attention to detect underwater targets. The innovations mainly include four aspects: Input, Backbone, Neck and Head, which include visual channel attention mechanism SE, random anchor box similarity calculation mechanism. We also replace the original Backbone with the two-stage cascade module of CSP(CSP2), which can extract features more comprehensively.

- A multi-stage fusion optimization scheme is proposed to improve performance. We design anchor boxes at 12 scales in UTD-Yolov5, and use WBF to fuse the filtered prediction boxes. After that, we iteratively refine them until the results are stable.

- Extensive experiments are designed to evaluate the UTD-Yolov5. We conduct experiments from multiple perspectives such as accuracy, recall, and computational speed. The experimental results show that UTD-Yolov5 can efficiently detect underwater targets, especially for complex coral reefs, which is far exceeds other baseline algorithms.

## II. RELATED WORK

Machine vision and deep learning are the main perception and control methods of robots, especially in the field of target detection. However, due to the weak underwater illumination, the images used for visual acquisition usually have problems such as low resolution and high noise, which previous vision-based methods difficult to be directly migrated. On the contrary, devices such as sonar, radar, etc. are widely used for robust perception of the underwater environment. Yu et al. [18] constructed a 3D sonar imaging system using sonar array cameras and multi-frequency acoustic signal emission for object recognition. In order to improve the perceptual effect of acoustic images, integral image methods [19], enhanced classifiers on Haar features [20] etc. are used to extract higher-order features. Devices such as laser scanners [21] and radars [22] have also been used in the study of underwater environment perception. However, studies have shown [23] that it is difficult to realize underwater target detection using ultrasound as a medium. The cost of these devices such as scanners is expensive, and the effect cannot be guaranteed under low light conditions.

Compared with acoustic perception, vision-based methods can achieve more efficient information collection at a lower cost. Aulinas [24] proposed a search method for salient color regions to obtain stable SURF features. Garcia [25] compared feature descriptors in high turbidity underwater images to obtain more complex information. Methods such as MSIS [26], probabilistic registration [27], and Cluster-based loop closing detection [28] can also achieve efficient underwater information collection. But these methods usually focus on partial information and are often used in the field of seafloor mapping or SLAM rather than object detection. The stereo matching system [29] and asynchronous stereo vision 3D reconstruction method [30] can provide more comprehensive information, but cannot meet the perceptual details requirements of object detection. Therefore, we decided to design a real-time object detection method that can target the features of underwater images with the recognition of complex CONTS and coral reefs as an entry point.

## III. MATERIALS AND METHODS

Our UTD-Yolov5 is a method for underwater object detection based on Attention Improved YOLOv5. It detects CONTS in coral reefs by performing high-level feature extraction and adaptive optimization on RGB images collected by underwater robots (such as AUV etc.). Different from the clear image quality in previous object detection, UTD-Yolov5 deals with lossy images with low light and low resolution. It realizes the framing of complex objects in high turbidity images through proposed methods such as multi-stage CSP2, SE, WBF, iterative refinement mechanism etc. Compared with the baseline methods, UTD-Yolov5 can distinguish the difference between CONTS and the surrounding highly similar environment more quickly, and eliminate the influence of underwater image quality. The algorithm framework of UTD-Yolov5 is shown in Figure 2.

We divide UTD-Yolov5 into two stages: the framework backbone Attention Improved YOLOv5 and multi-stage fusion optimization. We provide a detailed explanation of backbone innovations in the first part of this section. And for further optimization, we propose or introduce optimization methods such as WBF, iterative refinement, etc. This part will be introduced in the second stage.

### A. Attention Improved YOLOv5

Figure 2 shows the framework details of our UTD-Yolov5. By modularly replacing or cascading the Yolov5 network structure (covering 4 modules of the mainstream framework: input, backbone, neck and head.), we introduce CSP2, SE, etc. to achieve higher-order feature extraction. We also add a visual channel and spatial channel attention mechanism according to CBAM [31], and perform residual fusion based on weighted feature vectors. Finally, the location regression and category prediction of CONTS in coral reefs are realized.

YOLOv5 [1] is improved on the basis of the typical one-stage algorithm YOLOv4 [32]. YOLOv4 has achieved superior results on tasks on multiple online datasets such as COCO, ImageNet, etc. Among them, YOLOv4 can reach 43.5% AP and the speed is as high as 65 FPS on COCO. While YOLOv5 can achieve fast detection at 140 FPS on Tesla P100. Not only that, the volume of YOLOv5 is only 27 MB, which is nearly 90% smaller than that of YOLOv4 (244 MB) using the darknet architecture. In terms of accuracy indicators, the two are equivalent. The improvement points that greatly improve its speed and accuracy mainly include the following four aspects: a) Input: Adaptive anchor box calculation, adaptive image scaling, and Mosaic data enhancement. b) Backbone: CSP structure, Focus structure. c) Neck: The FPN and PAN structures are added between the backbone and head layers as a neck network. d) Head: loss function GIOU-Loss [1] during training and DIOU-nms [1] for prediction box screening.

According to the optimization ideas of YOLOv5, we innovate it from the following perspectives: a) weak illumination- make an enhanced classifier in the preprocessing and the Backbone to perform multi-scale segmentation and find differences in similar features; b)high turbidity- filter the input; c)the small size of the target(CONTS) and the high similarity with environment- designing structures to enable multi-scale similarity computation. Based on this, the specific improvements are as follows.

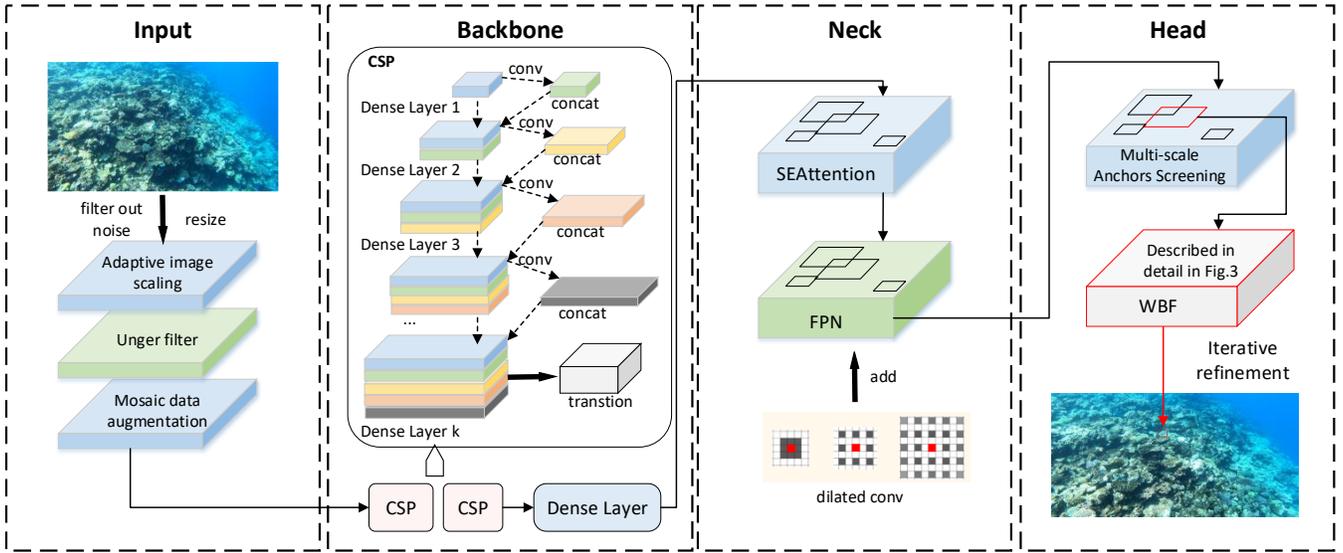

Figure 2. The framework of UTD-Yolov5.

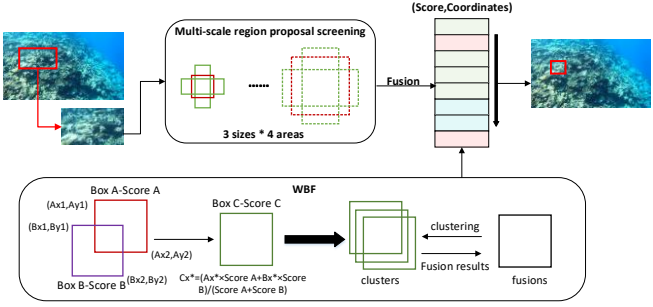

Figure 3. The framework of UTD-Yolov5.

**Input**. Before Mosaic data augmentation, Unger filter [33] is used to filter out noise in the input image. Next, we create a boundary to isolate seawater from coral reefs, and set a semi-automatic threshold to segment the results to prevent the effects of low light.

**Backbone**. Compared with the preprocessing part, which focuses on the quality of underwater images, we have made many improvements in the backbone network part to more efficiently extract the high-order features of the target under test. Due to the small size of CONTS itself and the complex environment of coral reefs, common object detection frameworks cannot clearly distinguish it from the surrounding environment. To more quickly delineate CONTS locations and segment features, we design a two-stage cascaded CSP network (CSP2), which can extract higher-order features and fuse them from multiple scales. The structural details are shown in Figure 2. In addition, the two-stage cascaded CSP2 network removes the convolution operation in the module connection, and is directly connected with a deep branch feature map. This operation can obtain more information without increasing the size of the model. This structure is widely used in our UTD-Yolov5.

**Neck**. In order to realize the feature segmentation of CONTS more flexibly, we introduce the visual channel attention mechanism module SE [34]. The SE network assigns weights to each channel through the three stages of Squeeze, Excitation, and Scale, which is showed in figure 2. in order to obtain a larger receptive field, we changed the last layer operation of YOLOv5's FPN network to dilated convolution [35]. We also design 12 anchor boxes with different specifications for more refined segmentation at this stage, and the details will be explained in the multi-stage fusion optimization section of this section.

**Head**. Different from YOLOv5's $Loss_{cls}$ and $Loss_{conf}$ for evaluating target and prediction boxes, we introduce random anchor box loss. This loss introduces a new reference metric for $CIoU$ by calculating the similarity between the predicted box and the random anchor box.

$$\alpha v = \frac{v^2}{(1-IoU)+\beta} \quad (1)$$

$$CIoU = IoU - \frac{l(rho^2)}{l(C^2)} - \frac{v^2}{(1-IoU)+\beta} \quad (2)$$

A good bounding box regressor contains three elements: Overlapping area, Central point distance and Aspect ratio. The parameters: *rho*- Euclidean distance between the center point of the prediction box and the GT box; *C*- Diagonal distance of the smallest closure region that contains both the prediction box and the GT box.

$$RAIoU = CIoU^* \{*: \text{random sampling}\} \quad (3)$$

$$L_{Loss} = L_{CIoU} + \sigma L_{RAIoU} \quad (4)$$

$RAIoU$ is the opposite value of the similarity between the background and the prediction frame after random sampling. $\sigma$ can be adjusted, [0,1]. $L_{Loss}$ is the loss function.

Experiments show that this design makes UTD-Yolov5 more accurate in the position prediction of CONTS, and the performance in class prediction is also improved to a certain extent.

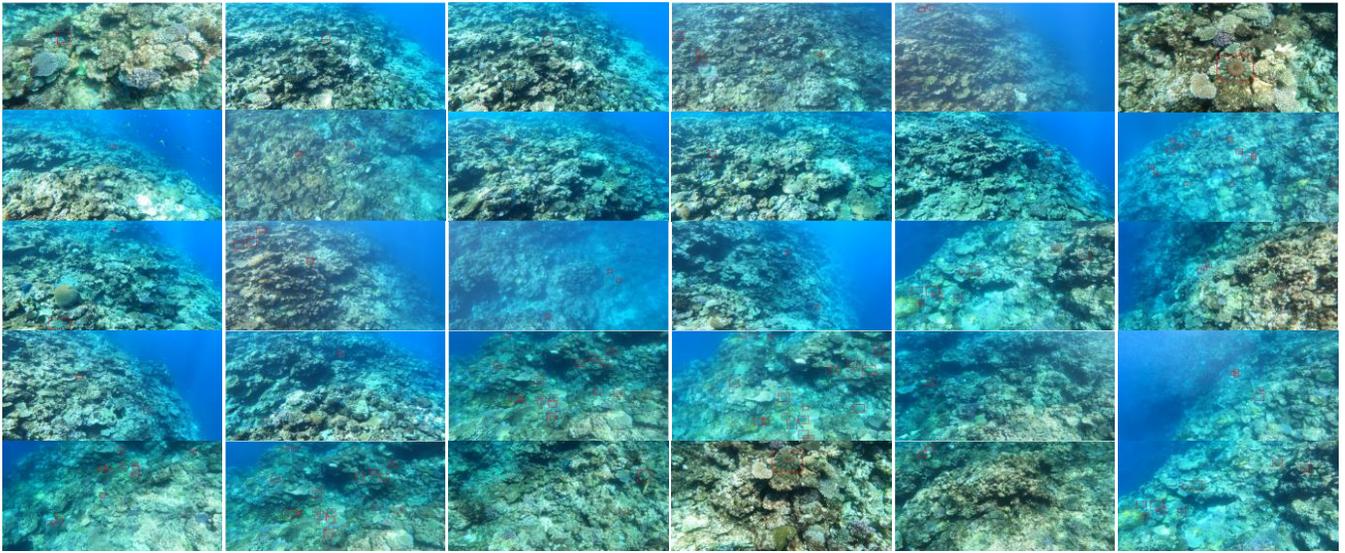

Figure 4. The results of UTD-Yolov5.

TABLE I. PERFORMANCE ON CSIRO

| Methods | AP | AP50 | AP75 | AR | ARs | ARm | ARl |
|---|---|---|---|---|---|---|---|
| Faster RCNN | 0.496 | 0.579 | 0.418 | 0.437 | 0.021 | 0.430 | 0.476 |
| FCOS | 0.527 | 0.684 | **0.515** | 0.571 | 0.035 | 0.519 | **0.668** |
| YOLOv3 | 0.398 | 0.503 | 0.406 | 0.421 | 0.017 | 0.398 | 0.458 |
| YOLOv4 | 0.415 | 0.622 | 0.511 | 0.487 | 0.023 | 0.401 | 0.542 |
| CornerNet | 0.330 | 0.491 | 0.351 | 0.513 | 0.004 | 0.513 | 0.610 |
| FPR | 0.512 | 0.610 | 0.466 | 0.495 | 0.027 | 0.474 | 0.602 |
| SNIPER | 0.349 | 0.673 | 0.417 | 0.414 | **0.036** | 0.426 | 0.551 |
| Fast RCNN+SGD | 0.460 | 0.597 | 0.395 | 0.536 | 0.031 | 0.501 | 0.631 |
| Fast RCNN+cyclic pooling | 0.472 | 0.631 | 0.526 | 0.548 | 0.041 | 0.528 | 0.629 |
| YOLOv5 | 0.502 | 0.549 | 0.511 | 0.491 | 0.028 | 0.477 | 0.596 |
| UTD-Yolov5 | **0.541** | **0.692** | 0.484 | **0.577** | 0.035 | **0.533** | 0.653 |

TABLE II. ABLATION STUDY

| | | | | | | |
|---|---|---|---|---|---|---|
| CSP2: 2-stage cascaded CSP | √ | √ | √ | √ | √ | √ |
| SE Attention | | √ | √ | √ | √ | √ |
| Dilated conv | | | √ | √ | √ | √ |
| Multiscale anchors screening | | | | √ | √ | √ |
| WBF | | | | | √ | √ |
| Iterative refinement | | | | | | √ |
| **Accuracy** | 69.4 | 70.5 | 71.7 | 74.8 | 75.3 | 78.5 |

### B. Multi-stage fusion optimization

Due to the particularity of underwater images and the complexity of the environment, we introduce a variety of optimization schemes to improve the performance of UTD-Yolov5.

In order to obtain more information to provide accurate position regression, we set up three sizes and four area proposal boxes. As shown in figure 3, we randomly crop detection frames with the same size and the same size outside the calibration frame, and calculate their similarity. This indicator also serves as a basis for separating CONTS from other marine organisms. After experiments, this specification design can achieve the optimal result in the link of generating the regional proposal frame.

In order to improve the calculation speed of UTD-Yolov5 and meet the purpose of real-time detection, we introduce the WBF algorithm to fuse and filter the generated multiple proposal boxes. Besides, we design an iterative refinement scheme to ensure that the obtained computational results are optimal. The upper limit of iteration $I$=3k. The entire optimization process is shown in Figure 3. Pseudocode is shown below.

## IV. EXPERIMENTAL RESULTS

In this section, we first introduce the dataset used for the experiments. UTD-Yolov5 is mainly tested on the CSIRO dataset. The second part provides a detailed description of the experimental design and results.

### A. Dataset

Our UTD-Yolov5 detects a particular species of starfish that threatens the Great Barrier Reef, the coral-eating crown of thorns, or COTS for short. The performance of the model is evaluated on an open source CONTS dataset, CSIRO [1].

This dataset is a large-scale annotated underwater image dataset comprising underwater image sequences collected at five different areas on coral reefs in the GBR. It contains over 35k images, hundreds of individual COTS. Each COTS of the training data is annotated with bounding boxes around it. Different from traditional datasets: a) the dataset contains only one category CONTS and is annotated based on sequences; b) the number of CONTS in the image to be detected is indeterminate and can overlap each other (given the habit of CONTS, there may be a large area of occlusion); c) due to the weak lighting and other characteristics of underwater images, there may be some high turbidity samples in the dataset.

*B. Experiments and Results*

Figure 4 shows the results of UTD-Yolov5 in underwater image detection. In addition to calibrating the final detection frame, the algorithm also prints the coordinates, length, width, and confidence of the detection frame. Table 1 shows the detection effects of various algorithms on CSIRO [1]. In order to comprehensively evaluate UTD-Yolov5, we report the two-stage algorithm represented by Faster RCNN [6] and the one-stage algorithm represented by YOLOv3 [7], including: Faster RCNN, YOLOv3, FCOS, CornerNet, etc. We also introduce fusion networks for testing, including: Faster RCNN+SGD fusion network, Faster RCNN+cyclic pooling, etc. In addition, in order to detect the improvement of the performance of YOLOv5 by our improvements to the network, YOLOv5 is also experimented as a comparison algorithm. The goal of this paper is to detect objects as comprehensively and quickly as possible, so recall is given priority.

To ensure objective results, all algorithms use the same segmentation mask. It can be seen from the table that the average detection accuracy of our UTD-Yolov5 in multiple experiments reaches 78.54%, which meets expectations and the robustness is verified. At the same time, it can be clearly seen that our proposed Attention Improved YOLOv5 has certain advantages compared with the original YOLOv5, and the performance has been improved.

We hope to have an in-depth and detailed analysis of the model to enhance interpretability and conduct ablation experiments. Due to the modular design of the entire network, we can activate or deactivate some modules at any time, such as: WBF, iterative refinement mechanism, visual attention module SE, preprocessing and so on. Table 2 shows the influence of different modules on the experimental results.

We can clearly find the impact of UTD-Yolov5's innovation in the optimization scheme on the accuracy, which improves it by nearly 7%. Our visual channel attention mechanism SE, CSP2, also improves the recall and accuracy of underwater target detection to varying degrees.

The CSP2 and other structures of UTD-Yolov5 are not only designed to improve the accuracy, but also hope to speed up the detection speed, thereby improving the working efficiency of the underwater robot. We set different time thresholds to explore the computational cost improvement of UTD-Yolov5 for underwater object detection. All algorithms are performed on the same computing platform. In experiments, we found that Multiscale anchors screening and WBF resulted in a significant increase in computational speed.

## V. Conclusion

In this paper, we propose an object detection network for underwater robots, called UTD-Yolov5. It takes the protection of coral reefs, a precious but fragile complex environment, as an entry point to do underwater detection. And CONTS is one of the biological threats, which is small in size, highly similar to the surrounding environment, and has different degrees of occlusion. These further increase the difficulty of detection and require the network to have strong high-order feature extraction capabilities. UTD-Yolov5 is improved by self-attention mechanism and multi-stage fusion based on YOLOv5. Our innovations are divided into two parts. The first stage proceeds from the 4 modules: input, backbone, neck and head, which include designing a two-stage cascaded CSP to replace the backbone (CSP2); introducing a visual channel attention mechanism SE module; adding a random anchor box similarity calculation mechanism, etc. In the second stage, considering the particularities of underwater images such as weak light and high turbidity, multi-stage fusion optimization is carried out. In this part, we propose iterative refinement mechanism and multi-scale random box selection, while use WBF to further improve the accuracy and speed. We have modularized the entire project to make it work more flexibly. Experiments show that UTD-Yolov5 achieves an average accuracy of 78.54% in underwater target detection, which meets expectations. Compared with the baseline algorithm, it has a great improvement in both accuracy and speed. This article takes the protection of coral reefs as the starting point. We will further explore various underwater environments in the future, and devote ourselves to the development of underwater robots.

References

[1] Liu, Jiajun, et al. "The CSIRO Crown-of-Thorn Starfish Detection Dataset." *arXiv preprint arXiv:2111.14311* (2021).

[2] Hoegh-Guldberg, Ove, et al. "Securing a long-term future for coral reefs." *Trends in Ecology & Evolution* 33.12 (2018): 936-944.

[3] Authority, Great Barrier Reef Marine Park. Great Barrier Reef Outlook Report 2019. *Great Barrier Reef Marine Park Authority*, 2019.

[4] Wilmes, Jennifer C., Andrew S. Hoey, and Morgan S. Pratchett. "Contrasting size and fate of juvenile crown-of-thorns starfish linked to ontogenetic diet shifts." *Proceedings of the Royal Society B* 287.1931 (2020): 20201052.

[5] Peng, Licheng, et al. "Micro-and nano-plastics in marine environment: Source, distribution and threats—A review." *Science of the Total Environment* 698 (2020): 134254.

[6] Chen, Xinlei, and Abhinav Gupta. "An implementation of faster rcnn with study for region sampling." *arXiv preprint arXiv:1702.02138* (2017).

[7] Redmon, Joseph, and Ali Farhadi. "Yolov3: An incremental improvement." *arXiv preprint arXiv:1804.02767* (2018).

[8] Singh, Bharat, Mahyar Najibi, and Larry S. Davis. "Sniper: Efficient multi-scale training." *Advances in neural information processing systems* 31 (2018).

[9] Law, Hei, and Jia Deng. "Cornernet: Detecting objects as paired keypoints." *Proceedings of the European conference on computer vision (ECCV)*. 2018.

[10] Zhou, Xingyi, Jiacheng Zhuo, and Philipp Krahenbuhl. "Bottom-up object detection by grouping extreme and center points." *Proceedings of the IEEE/CVF conference on computer vision and pattern recognition*. 2019.

[11] Paz, David, Hengyuan Zhang, and Henrik I. Christensen. "Tridentnet: A conditional generative model for dynamic trajectory generation." *arXiv preprint arXiv:2101.06374* (2021).

[12] Tian, Zhi, et al. "Fcos: Fully convolutional one-stage object detection." *Proceedings of the IEEE/CVF international conference on computer vision*. 2019.

[13] Guo, Dong, et al. "Correcting over-exposure in photographs." *2010 IEEE Computer Society Conference on Computer Vision and Pattern Recognition*. IEEE, 2010.

[14] Shrivastava, Abhinav, and Abhinav Gupta. "Contextual priming and feedback for faster r-cnn." *European conference on computer vision. Springer*, Cham, 2016.

[15] Jiang, Huaizu, and Erik Learned-Miller. "Face detection with the faster R-CNN." *2017 12th IEEE international conference on automatic face & gesture recognition (FG 2017)*. IEEE, 2017.

[16] Zhang, Junkang, et al. "Cancer cells detection in phase-contrast microscopy images based on Faster R-CNN." *2016 9th international symposium on computational intelligence and design (ISCID)*. Vol. 1. IEEE, 2016.

[17] Zhang, Qi-xing, et al. "Wildland forest fire smoke detection based on faster R-CNN using synthetic smoke images." *Procedia engineering* 211 (2018): 441-446.

[18] Yu, Son-Cheol, Zhu Teng, and Dong-Joong Kang. "Modeling of high-resolution 3d sonar for image recognition." *International Journal of Offshore and Polar Engineering* 22.03 (2012).

[19] D.P. Williams and J. Groen. "A Fast Physics-Based, Environmentally Adaptive Underwater Object Detection Algorithm". *In Proc. of OCEANS*, pages 1–7, 2011.

[20] J. Sawas, Y.R. Petillot, and Y. Pailhas. "Cascade of Boosted Classifiers for Rapid Detection of Under- water Objects". *In ECUA 2010 Istanbul Conference*, pages 1–8, 2010.

[21] Mertz, Christoph, et al. "Moving object detection with laser scanners." *Journal of Field Robotics* 30.1 (2013): 17-43.

[22] Nabati, Ramin, and Hairong Qi. "Rrpn: Radar region proposal network for object detection in autonomous vehicles." *2019 IEEE International Conference on Image Processing (ICIP)*. IEEE, 2019.

[23] P. Jonsson, I. Sillitoe, B. Dushaw, J. Nystuen, and J. Heltne. "Observing using sound and light: a short review of underwater acoustic and video-based methods". *Ocean Science Discussions*, 6(1):819–870, 2009.

[24] J. Aulinas, M. Carreras, X. Llado, J. Salvi, R. Garcia, R. Prados, and Y.R. Petillot. "Feature extraction for underwater visual SLAM". *In TODO: FIX*, pages 1–7, 2011.

[25] R. Garcia and N. Gracias. "Detection of interest points in turbid underwater images". *In IEEE OCEANS*, pages 1–9, 2011.

[26] Wawrzyniak, Natalia, and Grzegorz Zaniewicz. "Detecting small moving underwater objects using scanning sonar in waterside surveillance and complex security solutions." *2016 17th International Radar Symposium (IRS)*. IEEE, 2016.

[27] Tsuzuki, Daisuke, et al. "Stable and convenient spatial registration of stand-alone NIRS data through anchor-based probabilistic registration." *Neuroscience Research* 72.2 (2012): 163-171.

[28] Negre, Pep Lluis, Francisco Bonin-Font, and Gabriel Oliver. "Cluster-based loop closing detection for underwater slam in feature-poor regions." *2016 IEEE International Conference on Robotics and Automation (ICRA)*. IEEE, 2016.

[29] J.P. Queiroz-Neto, R. Carceroni, W. Barros, and M. Campos. Underwater stereo. In Computer Graphics and Image Processing, 2004. *Proceedings. 17th Brazilian Symposium on*, pages 170–177, 2004.

[30] A. Leone, G. Diraco, and C. Distante. Stereoscopic system for 3-d seabed mosaic reconstruction. volume 2, pages 541–544, Sep. 2007.

[31] Woo, Sanghyun, et al. "Cbam: Convolutional block attention module." *Proceedings of the European conference on computer vision (ECCV)*. 2018.

[32] Bochkovskiy, Alexey, Chien-Yao Wang, and Hong-Yuan Mark Liao. "Yolov4: Optimal speed and accuracy of object detection." *arXiv preprint arXiv:2004.10934* (2020).

[33] de Cheveigné, Alain, and Israel Nelken. "Filters: when, why, and how (not) to use them." *Neuron* 102.2 (2019): 280-293.

[34] Fukui, Hiroshi, et al. "Attention branch network: Learning of attention mechanism for visual explanation." *Proceedings of the IEEE/CVF conference on computer vision and pattern recognition*. 2019.

[35] Zhang, Xiaohu, Yuexian Zou, and Wei Shi. "Dilated convolution neural network with LeakyReLU for environmental sound classification." *2017 22nd international conference on digital signal processing (DSP)*. IEEE, 2017.